\documentclass[cameraready]{Interspeech}
\usepackage{multirow}
\usepackage{booktabs}  
\usepackage{graphicx}
\usepackage{float}
\usepackage{pifont}  
\newcommand{\circled}[1]{\textcircled{\raisebox{-0.9pt}{\small #1}}}

\title{Automated Gradient-Driven Parameter Sharing for Low-Resource Multilingual Speech-to-Text Translation}
\author[affiliation={1}]{Ruiyan}{Sun}
\author[affiliation={2}]{Satoshi}{Nakamura}

\address{
    $^1$ School of Data Science \\
    $^2$ School of Aritificial Intelligence,  The Chinese University of Hong Kong,Shenzhen
}

\email{ruiyansun@link.cuhk.edu.cn}


\usepackage{amsmath,amssymb,amsfonts}
\usepackage{tabularx}
\usepackage{array}

\usepackage{comment}

\begin{document}

\maketitle

\begin{abstract}
    In low-resource multilingual speech-to-text translation, uniform architectural sharing across languages frequently introduces representation conflicts that impede convergence. This work proposes a principled methodology to automatically determine layer-specific sharing patterns by mining training gradient information. Our approach employs three distinct analysis strategies: distance-based language clustering, self/cross-task divergence metrics for capacity allocation, and joint factorization coupled with canonical correlation analysis for subspace alignment. Extensive evaluation across four language pairs (using the SeamlessM4T-Medium architecture) demonstrates persistent improvements in translation quality metrics.
\end{abstract}

\section{Introduction}

Building effective speech-to-text systems for multiple languages with limited training resources remains a persistent challenge. A core architectural decision involves how parameters are distributed across languages: rigid uniform sharing frequently fails to account for linguistic diversity, while language-specific models suffer from sparse data and weak cross-lingual transfer. The tension between shared and specialized components necessitates principled design approaches.

Gradient conflict is recognized as a primary bottleneck in multilingual learning. Optimization techniques such as PCGrad~\cite{yu2020gradient}, GradOPS~\cite{zhu2025gradops}, and GDOD~\cite{dong2022gdod} stabilize training via orthogonal gradient projections, while structural strategies like shared-private representations~\cite{bousmalis2016domain} and task disentangling~\cite{yang2026disentangling} seek to decouple interfering parameters. Our specialization of FFN2 in Conformer is motivated by high parameter density and role in non-linear feature transformation~\cite{gerber2025attention}, rendering them more malleable for language-specific features than attention modules.

Despite these advances, a significant gap remains: designing optimal parameter sharing configurations is prohibitively high. Most existing "shared-private" or "expert squads" architectures depend on human intuition or expensive neural architecture search (NAS) to determine layer-wise configurations.

We present an automated framework that derives optimal architectural specialization by analyzing optimization dynamics across languages. The pipeline consists of three phases: \textit{Training Dynamic Analysis $\rightarrow$ Architecture Configuration $\rightarrow$ Specialized Fine-tuning}. By integrating clustering, similarity-based metrics, and subspace analysis, the approach automatically identifies language groupings and decomposition parameters without manual intervention.

The method addresses three core challenges: \textbf{linguistic heterogeneity}, where undifferentiated sharing causes interference; \textbf{data scarcity}, which amplifies optimization noise; and \textbf{architectural scalability}, which makes manual design infeasible. Our contributions include: (i) a systematic analysis framework connecting training dynamics to architectural decisions; (ii) instantiation within a standard speech-to-text backbone, targeting high-density transformer blocks; (iii) empirical demonstration of consistent improvements across multiple evaluation metrics on four language pairs. \footnote{Code and additional resources will be released following review completion.}

\section{Data}

Experiments utilize data from the IWSLT 2025 Low-resource Speech-to-Text track. We sub-sample the larger aeb, bem, and est corpora to ensure a balanced distribution for gradient analysis relative to the smaller gle dataset. All speech inputs are preprocessed at 16kHz following standard IWSLT 2025 protocols. Datasets are obtained from official IWSLT 2025 tracks with train/validation/test splits in approximately 10:1:1 ratio, as summarized in Table~\ref{tab:datasets}.

\begin{table}[th]
  \caption{Dataset summary for 2-way ST translated to English.}
  \label{tab:datasets}
  \centering
  \resizebox{\columnwidth}{!}{%
  \begin{tabular}{lccl}
    \toprule
    \textbf{Language} & \textbf{Task} & \textbf{Amount} & \textbf{Sources} \\
    \midrule
    Tunisian (aeb) & 2-way ST & 20k lines & IWSLT2022~\cite{anastasopoulos2022findings} \\
    Bemba (bem)    & 2-way ST & 20k lines & BIG-C~\cite{sikasote2023bigc} \\
    Estonian (est) & 2-way ST & 20k lines & LoResMT~\cite{sildam2024finetuning} \\
    Irish (gle)    & 2-way ST & 7k lines  & IWSLT2023~\cite{agarwal2023findings} \\
    \bottomrule
  \end{tabular}%
  }
\end{table}

\section{Methodology}

\subsection{Framework Construction}

\begin{figure}[H]
  \centering
  \includegraphics[width=\linewidth]{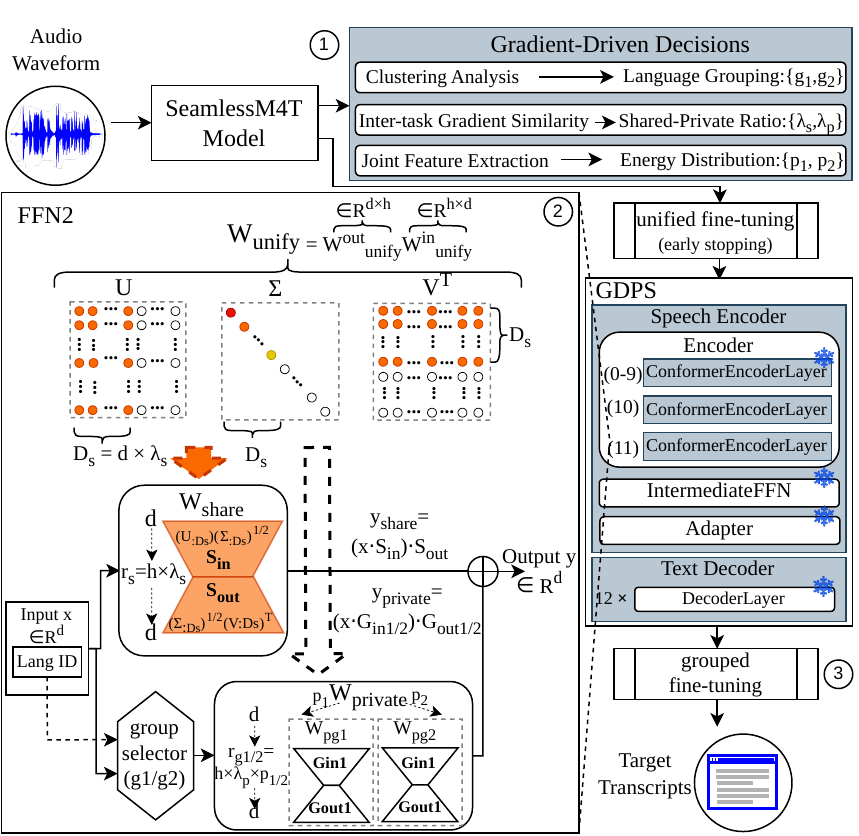}
  \caption{Overview of the GDPS framework combining the incorporation of gradient-driven decision-making, dynamic parameter configuration, and fine-tuning.}
  \label{fig:architecture}
\end{figure}

As illustrated in Figure~\ref{fig:architecture}, the GDPS framework consists of three key components: \circled{1} \textbf{Gradient-Driven Decision-Making} analyzes inter-language gradient behaviors to automatically determine optimal language groupings and shared-private parameter ratios (detailed in Section~\ref{subsec:gradient_analysis}); \circled{2} \textbf{Dynamic Parameter Configuration} instantiates these decisions by decomposing Encoder Layer 11 FFN2 into shared and language-specific private branches (detailed in Section~\ref{subsec:architecture_design}); \circled{3} \textbf{Grouped Fine-tuning} optimizes the specialized GDPS architecture with group-wise parameter updates. This unified design eliminates manual architectural search while preserving cross-lingual transfer capacity.

\subsection{Gradient Analysis Formulations}\label{subsec:gradient_analysis}

Our framework integrates three complementary techniques to bridge training dynamics and structural design.

\subsubsection{Method A: Language Grouping via Clustering}

We first compute the pairwise gradient cosine similarity between languages at specific layers:
\begin{equation} 
s_{i,j} = \frac{g_i \cdot g_j}{\Vert g_i \Vert \Vert g_j \Vert} 
\label{eq:sim} 
\end{equation}
This produces a similarity matrix $S$:
\begin{equation}
S = \begin{bmatrix} 1 & s_{1,2} & \dots & s_{1,n} \\ s_{2,1} & 1 & \dots & s_{2,n} \\ \vdots & \vdots & \ddots & \vdots \\ s_{n,1} & s_{n,2} & \dots & 1 \end{bmatrix}
\label{eq:sim_matrix}
\end{equation}
where $g_i$ and $g_j$ denote the \textit{averaged} gradients of language $i$ and $j$. We use this \textbf{language-level similarity} to capture the macro-directional orientation of each language, which is more robust for identifying stable grouping clusters. We convert similarity to distance using $d_{i,j} = 1 - s_{i,j}$. K-means clustering is applied to minimize intra-cluster variance:
\begin{equation}
\min_{\{C_k\}} \sum_{k=1}^{K} \sum_{i \in C_k} \Vert x_{i} - \mu_k \Vert^2
\label{eq:kmeans}
\end{equation}
where $\mu_k$ is the centroid of cluster $C_k$. Hierarchical clustering iteratively merges clusters using single linkage distance:
\begin{equation}
d(C_i, C_j) = \min_{a \in C_i, b \in C_j} d(a, b)
\label{eq:hierarchical}
\end{equation}
This identifies language grouping structures where languages in the same cluster share parameters.

\subsubsection{Method B: Self versus Cross Gradient Similarity}

We analyze candidate bottleneck regions by defining self-task similarity ($S_{\text{self}}$) and cross-task similarity ($S_{\text{cross}}$). \textbf{Self-task} similarity measures the average gradient alignment between different samples within the same language task, and \textbf{cross-task} similarity captures the alignment across distinct translation directions. These metrics allow us to distinguish between task-specific noise and cross-lingual signal overlap:
\begin{align}
S_{\text{self}} &= \mathbb{E}_t \mathbb{E}_{i \neq j} \cos(g_{t,i}, g_{t,j}) \\
S_{\text{cross}} &= \mathbb{E}_{t \neq t'} \mathbb{E}_{i,j} \cos(g_{t,i}, g_{t',j})
\end{align}
where $g_{t,i}$ is the gradient of the $i$-th \textit{individual sample} in language $t$. This \textbf{sample-level similarity} quantifies the precise degree of pairwise conflict and distribution overlap between tasks. The conflict strength scalar $\delta$ is defined as the mean difference $S_{\text{self}} - S_{\text{cross}}$ between the candidate layers. We map $\delta$ to the shared ratio using a piecewise function\footnote{The sensitivity of these thresholds is discussed in the ablation study in Section 4.5.}:
\begin{equation}
\text{SharedRatio} = 
\begin{cases} 
0.75 & \delta < 0.05 \\
0.50 & 0.05 \leq \delta < 0.15 \\
0.25 & \delta \geq 0.15 
\end{cases}
\end{equation}

\subsubsection{Method C: Joint SVD and Regularized CCA}

Let $G_i \in \mathbb{R}^{m \times d}$ be the gradient matrix of language $i$. We perform Joint SVD on the concatenated matrix:
\begin{equation}
G_{\text{concat}} = [G_1; G_2; \dots; G_n] = U \Sigma V^\top
\label{eq:svd}
\end{equation}
We project gradients onto the top-$k$ subspace defined by the first $k$ right singular vectors $V_k$. To capture the linear alignment between language-specific gradient subspaces, we follow the ridge-regularized canonical correlation analysis (CCA) formulation~\cite{hotelling1936relations, hardoon2004canonical}. We maximize the normalized cross-covariance between languages $i$ and $j$:
\begin{align}
\rho_{ij} &= \max_{w_i, w_j} \frac{w_i^\top \mathbf{\Gamma}_{ij} w_j}{\sqrt{D_{w}}} \label{eq:cca} \\
D_{w} &= (w_i^\top (\mathbf{\Gamma}_{ii} + \lambda I) w_i)(w_j^\top (\mathbf{\Gamma}_{jj} + \lambda I) w_j) \nonumber
\end{align}
where $\mathbf{\Gamma}_{ij}$ and $\mathbf{\Gamma}_{ii}$ denote the cross-covariance and auto-covariance matrices. The denominator $D_w$ normalizes by the product of projected variances, enabling scale-invariant correlation maximization. The projection vectors $w_i, w_j$ identify directions of maximal linear correlation. Ridge regularization with penalty term $\lambda I$~\cite{hardoon2004canonical, tuzhilina2023canonical} ensures robust covariance estimation in high-dimensional settings. The leading canonical correlation magnitude reflects task subspace alignment, which we use to calibrate private module initialization~\cite{andrew2013deep}. 

The energy $E_i$ captured by these principal directions is:
\begin{equation}
E_i = \sum_{j=1}^k \Vert G_i v_j \Vert^2
\label{eq:energy}
\end{equation}
The energy proportion $p_i$ for language $i$ is defined as $p_i = E_i / \sum_{l=1}^n E_l$.

\subsection{Gradient Observations and Analysis Results}

Applying our empirical framework to the SeamlessM4T-Med backbone yields several critical insights that directly inform and validate our subsequent architectural configuration choices.

\textbf{Layer Selection and the Purity Paradox:} Our layer-wise evaluation reveals a fundamental trade-off. Specifically, Layer 11 exhibits much stronger gradient signal strength ($S_{\text{self}} \approx 0.42$--$0.51$ compared to $0.22$--$0.30$ observed in L10) but simultaneously experiences a significant $6.6\%$--$12.6\%$ purity decline. This systematic trade-off consistently limits optimization across all language pairs, empirically isolating Layer 11 FFN2 as the primary network bottleneck where severe cross-lingual representation conflicts intersect during training.

\textbf{Language Grouping (Method A):} Distance-based clustering robustly partitions the target languages into Group 1 (Bem) and Group 2 (Aeb, Est, Gle). Our computed distance matrix strongly supports this division, showing that Bemba remains highly isolated in optimization space ($0.243$ distance to Gle). Conversely, Aeb and Est demonstrate high proximity ($0.157$), indicating they naturally form a compatible representation subspace suitable for shared parameterization.

\textbf{Shared-Private Ratio (Method B):} We observe average conflict scores of $\delta \approx 0.075$ (with an empirical margin of $0.07$--$0.08$). When combined with the aforementioned purity degradation in Layer 11, this score confirms a substantial structural gradient divergence. By mapping this value precisely via Eq. 7, we derive and adopt a strict 50\% shared parameter ratio, which provides an optimal structural balance between maximizing cross-lingual knowledge transfer and allowing sufficient task-specific specialization.

\textbf{Energy Distribution (Method C):} Thorough Joint SVD evaluation of the aggregated gradients reveals that the very first principal component consistently captures $\approx 55\%$ of the total gradient energy. Furthermore, the calculated Gini coefficients remain notably high ($0.63$--$0.72$). This directly confirms a massive concentration of critical optimization information within the top-$k$ singular vectors, robustly validating our mathematical decision to leverage these specific principal directions for targeted, task-relevant module initialization.

\subsection{GDPS Framework Design and Implementation}\label{subsec:architecture_design}

Based on the gradient analysis results in Section 3.3, we instantiate the GDPS framework by specializing the FFN2 layer of Encoder Layer 11 with three key design components.

\textbf{Specialized Routing (Grouping):} Based on clustering, tokens from Bemba are routed through  Group 1, while tokens from Aeb, Est, and Gle are routed through  Group 2.

\textbf{Dimension Splitting (Ratio):} Following the 50\% ratio found via Conflict Score $\delta$, we decompose the FFN2 weight $\mathbf{W}_{\text{unified}} = \mathbf{W}_2 \mathbf{W}_1$ (where $\mathbf{W}_1 \in \mathbb{R}^{d \times 1024}$ and $\mathbf{W}_2 \in \mathbb{R}^{1024 \times d}$ with $d=4096$). Define the equivalent low-rank factorization $\mathbf{W}_{\text{equiv}} = \mathbf{W}_2 \mathbf{W}_1 \in \mathbb{R}^{1024 \times 1024}$ and compute its SVD: $\mathbf{W}_{\text{equiv}} = \mathbf{U} \mathbf{\Sigma} \mathbf{V}^\top$. We derive symmetric factors:
\begin{align}
\mathbf{W}_{1,\text{factor}} &= \mathbf{U}_{:,:D_s/4} \sqrt{\mathbf{\Sigma}_{:D_s/4,:D_s/4}} \quad (1024 \times D_s/4) \\
\mathbf{W}_{2,\text{factor}} &= \sqrt{\mathbf{\Sigma}_{:D_s/4,:D_s/4}} \mathbf{V}_{:,:D_s/4}^\top \quad (D_s/4 \times 1024)
\end{align}
where $D_s = 2048$ is the target shared dimensionality. These factors are expanded to $D_s$ by padding with small Gaussian noise and transpose to match FFN layer shapes. The private modules for each group $g$ split the residual capacity $D_p = (4096 - D_s) / N = 1024$ per group. The final output is:
\begin{equation}
    \text{FFN2}(x) = \text{Activation}(x \mathbf{W}_s) \oplus \text{Activation}(x \mathbf{W}_{pg})
\end{equation}

\textbf{Energy-Driven Residual Initialization:} To initialize private modules with residual knowledge, we first reconstruct the shared-only factorization $\mathbf{W}_{\text{shared,equiv}} = \mathbf{W}_{2,\text{factor}} \mathbf{W}_{1,\text{factor}}$ and compute the residual:
\begin{equation}
\mathbf{W}_{\text{res}} = \mathbf{W}_{\text{equiv}} - \mathbf{W}_{\text{shared,equiv}}
\end{equation}
We allocate this residual to each language group $g$ proportionally to their gradient energy $p_g$ from Method C. For each group, we perform SVD on the energy-weighted residual:
\begin{equation}
\mathbf{W}_{\text{res},g} = p_g \mathbf{W}_{\text{res}}
\end{equation}
and extract the top-$D_p / N = 1024$ components (for $N=2$ groups), then expand to $D_p = 1024$ dimensions for initializing $\mathbf{W}_{pg}$. This energy-weighted initialization ensures groups with higher gradient energy receive stronger residual patterns while preventing cold-start failures and resolving identified representation conflicts.

\section{Experiments and results}

\subsection{Experimental Setup and Training Configuration}

We use SeamlessM4T-Medium\footnote{https://huggingface.co/facebook/hf-seamless-m4t-medium} as the backbone model (1.2B parameters, 12 conformer encoder and decoder layers) based on the official seamless\_communication framework. All experiments are conducted on NVIDIA A100 GPUs with mixed-precision training (FP16) using the AdamW optimizer. Training hyperparameters are configured as follows: base learning rate $\alpha = 4\times 10^{-5}$ (group-adjusted: $\alpha_g = 10^{-4}$), batch size $B = 4$, dropout $p_d = 0.05$, weight decay $\lambda = 0.05$, random seed $s = 2343$, and warmup steps $t_w = 2000$.

\subsection{Main Results}

Table~\ref{tab:main_results} compares the proposed \textbf{GDPS} architecture with SeamlessM4T-Medium baseline and Unified fine-tuning. Our GDPS configuration, which integrates three gradient-driven techniques, achieves consistent improvements in BLEU, TER, BERTScore, and COMET scores across all languages. Specifically, we observe relative COMET gains of up to 3.26\% over the Unified FT baseline, validating that gradient-driven architectural specialization effectively mitigates inter-task interference while preserving the universal shared subspace.

\begin{table}[th]
  \caption{Baseline vs. Unified fine-tuning vs. GDPS Architecture on SeamlessM4T-Medium.}
  \label{tab:main_results}
  \centering
  \resizebox{\columnwidth}{!}{%
  \begin{tabular}{llcccc}
    \toprule
    \textbf{Pairs} & \textbf{Method} & \textbf{BLEU $\uparrow$} & \textbf{TER $\downarrow$} & \textbf{BERT $\uparrow$} & \textbf{COMET $\uparrow$} \\
    \midrule
    \multirow{3}{*}{Aeb-en} & SeamlessM4T-Med & 3.19 & 96.33 & 0.1635 & 0.5017 \\
                            & Unified FT & 7.64 & 97.17 & 0.2453 & 0.5326 \\
                            & \textbf{GDPS (Ours)} & \textbf{8.74} & \textbf{92.36} & \textbf{0.2745} & \textbf{0.5500} \\
    \midrule
    \multirow{3}{*}{Bem-en} & SeamlessM4T-Med & 0.82 & 122.82 & 0.1065 & 0.4037 \\
                            & Unified FT & 18.45 & 74.74 & 0.5252 & 0.6866 \\
                            & \textbf{GDPS (Ours)} & \textbf{19.69} & \textbf{73.50} & \textbf{0.5388} & \textbf{0.7012} \\
    \midrule
    \multirow{3}{*}{Est-en} & SeamlessM4T-Med & 11.29 & 69.44 & 0.5215 & 0.7102 \\
                            & Unified FT & \textbf{16.68} & 62.91 & 0.5902 & 0.7363 \\
                            & \textbf{GDPS (Ours)} & 16.49 & \textbf{62.36} & \textbf{0.5953} & \textbf{0.7414} \\
    \midrule
    \multirow{3}{*}{Gle-en} & SeamlessM4T-Med & 30.62 & 66.87 & 0.4371 & 0.6620 \\
                            & Unified FT & 43.59 & 51.08 & 0.5701 & 0.7257 \\
                            & \textbf{GDPS (Ours)} & \textbf{46.20} & \textbf{48.19} & \textbf{0.5959} & \textbf{0.7473} \\
    \bottomrule
  \end{tabular}%
  }
\end{table}

\subsection{Competitive Comparison with IWSLT Systems}

We evaluate \textbf{GDPS} against IWSLT benchmarks~\cite{robinson2025jhu} following the official low-resource protocol. While the current SOTA (GMU~\cite{meng2025gmu}) achieves superior scores by leveraging massive auxiliary datasets (and is omitted from Table~\ref{tab:competition} for a more commensurate comparison), GDPS enhances cross-lingual transfer efficiency via gradient-aligned parameter sharing, achieving robust performance under data-constrained regimes.

\begin{table}[th]
  \caption{Comparison with IWSLT 2025 Benchmarks (Eval BLEU).}
  \label{tab:competition}
  \centering
  \resizebox{\columnwidth}{!}{%
  \begin{tabular}{lcccc}
    \toprule
    \textbf{Method} & \textbf{Aeb-en} & \textbf{Bem-en} & \textbf{Est-en} & \textbf{Gle-en} \\
    \midrule
    Seamless v2~\cite{robinson2025jhu} & 6.70 & 14.67$^*$ & - & 12.30 \\
    JHU (MBR-BLEU)~\cite{robinson2025jhu} & 8.20 & \textbf{26.80} & - & 11.60 \\
    \midrule
    \textbf{GDPS (Ours)} & \textbf{8.39} & 20.29 & \textbf{20.30} & \textbf{40.20} \\
    \bottomrule
  \end{tabular}%
  }

  \smallskip
  \footnotesize{$^*$Note: Bem-en Eval result not available; Test1 used. $^\dagger$Indicates the overall SOTA system (IWSLT 2025 winner) in most tracks.}
\end{table}

Under strict low-resource constraints, these results demonstrate that GDPS's gradient-aligned biases can effectively catalyze cross-lingual transfer without requiring the extensive external corpora used by larger-scale systems.

\subsection{Gradient Representation Alignment Analysis}

\begin{figure}[!ht]
  \centering
  \includegraphics[width=\linewidth]{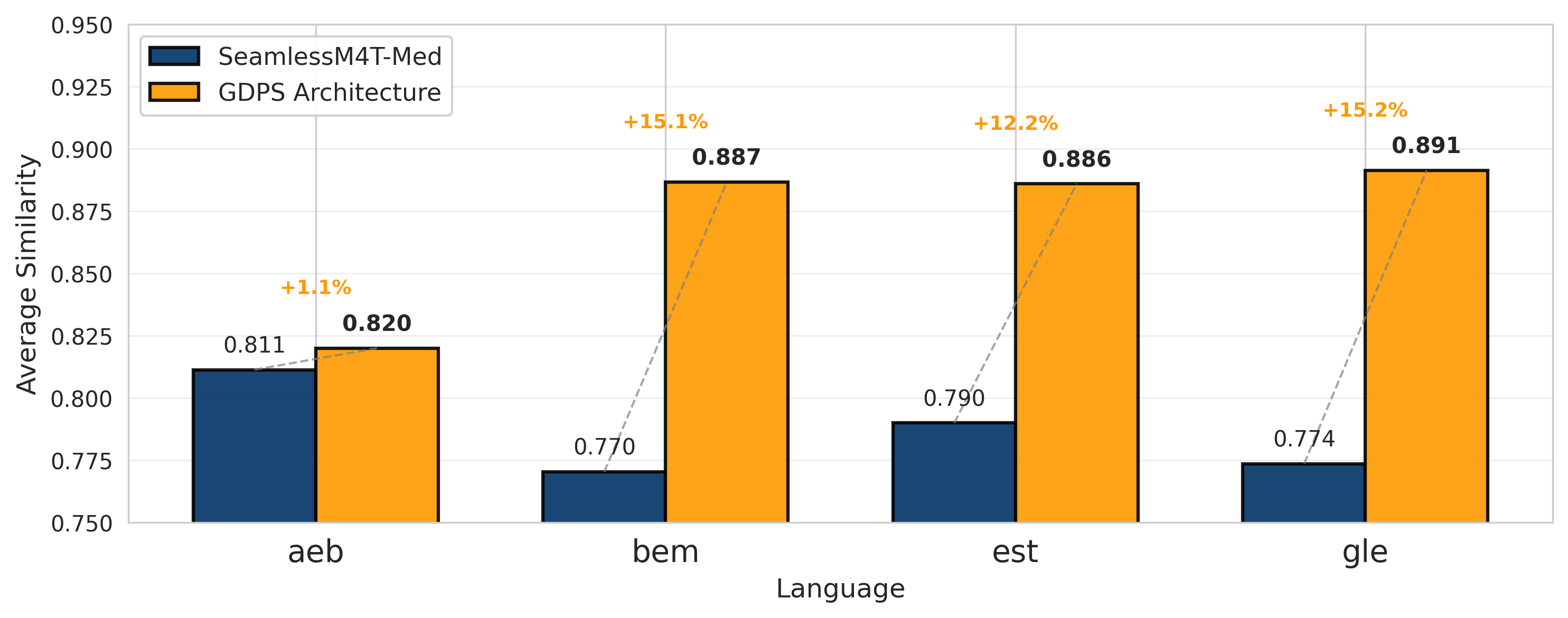}
  \caption{Average cross-language similarity per language between GDPS and the SeamlessM4T-Med baseline.}
  \label{fig:gradient_similarity}
\end{figure}

GDPS significantly enhances cross-lingual alignment compared to the SeamlessM4T-Med baseline. As shown in Figure~\ref{fig:gradient_similarity}, average similarity increases across all languages, with especially notable gains for Irish (+15.2\%) and Bemba (+15.1\%), indicating that shared-private decomposition effectively isolates task interference. Overall, GDPS achieves a +0.085 average similarity increase, validating that automated grouping and initialization facilitate efficient knowledge transfer.

\subsection{Ablation Analysis}

\begin{table}[th]
  \caption{Consolidated Ablation Results. Metrics are reported within each cell as \textbf{BLEU / COMET}. GDPS (Full) denotes the complete framework serving as the baseline for all variations.}
  \label{tab:ablation_master}
  \centering
  \resizebox{\columnwidth}{!}{%
  \begin{tabular}{lcccc}
    \toprule
    \textbf{Setting} & \textbf{Aeb-en} & \textbf{Bem-en} & \textbf{Est-en} & \textbf{Gle-en} \\
    \midrule
    \textbf{GDPS(Full)} & \textbf{8.74 / 0.5500} & \textbf{19.69 / 0.7012} & \textbf{16.49 / 0.7414} & \textbf{46.20 / 0.7473} \\
    \midrule
    \textit{Components} \\
    - Only A & 7.90 / 0.5381 & 19.63 / 0.6937 & 15.70 / 0.7374 & 42.85 / 0.7264 \\
    - Only B & 8.20 / 0.5479 & 19.22 / 0.6973 & 16.27 / 0.7368 & \textbf{47.01} / 0.7449 \\
    - Only C & 7.87 / 0.5348 & 18.93 / 0.6945 & 16.24 / 0.7376 & 43.77 / 0.7311 \\
    \midrule
    \textit{Ratio ($\delta$)} \\
    - 75\% shared & 8.22 / 0.5449 & 19.31 / 0.6980 & 16.53 / 0.7390 & \textbf{47.27} / 0.7448 \\
    - 25\% shared & 7.86 / 0.5381 & 19.17 / 0.6963 & \textbf{16.64} / 0.7358 & 43.69 / 0.7354 \\
    \midrule
    \textit{SVD Top-k} \\
    - k = 20 & \textbf{8.98 / 0.5542} & 19.36 / 0.6971 & 16.09 / 0.7386 & \textbf{46.93 / 0.7513} \\
    - k = 5 & 7.57 / 0.5401 & 18.89 / 0.6971 & \textbf{16.89} / 0.7399 & 45.46 / 0.7408 \\
    \midrule
    \textit{Data Scale} \\
    - 50\% Data & 8.07 / 0.5373 & 18.43 / 0.6899 & 15.96 / 0.7332 & 42.42 / 0.7187 \\
    - 66\% Data & 7.95 / 0.5454 & 18.63 / 0.6905 & 15.66 / 0.7392 & 41.81 / 0.7261 \\
    \midrule
    \textit{Location/Efficacy} \\
    - L11 FFN1 & 8.40 / 0.5434 & 19.13 / 0.6986 & 16.13 / 0.7352 & 41.10 / 0.7238 \\
    - L10 FFN2 & 8.26 / 0.5463 & 18.57 / 0.6968 & \textbf{16.53} / 0.7381 & \textbf{46.55} / 0.7465 \\
    - Adapter FFN & 6.24 / 0.5136 & 15.98 / 0.6565 & 15.23 / 0.7307 & 41.38 / 0.7077 \\
    \bottomrule
  \end{tabular}%
  }
\end{table}

Table~\ref{tab:ablation_master} consolidates sensitivity studies across analysis components, conflict thresholds, subspace dimensionality, data scale, and module localization.

Ablations reveal that removing any component (Methods A, B, or C) causes decay, confirming synergy. The adoption of 50\% shared ratio, derived from our $\delta$ thresholding, emerges as the optimal configuration for this language set; alternative ratios (25\%, 75\%) lead to suboptimal trade-offs between transfer and interference, validating our threshold design. Performance scales with data, indicating GDPS serves as a stable inductive bias. Importantly, applying GDPS to low-conflict modules (L10 FFN2, L11 FFN1) or the Subsampling Adapter results in marginal gains or degradation. This suggests forcing specialization in general-representation modules may impair generalization, confirming that GDPS efficacy depends on accurate conflict-driven localization.

\section{Conclusions}

We presented a systematic methodology for automated architectural design in low-resource multilingual speech translation. Through integration of gradient-based analysis techniques, the proposed approach achieves consistent performance improvements over baseline fine-tuning strategies. Our findings demonstrate that layer-wise parameter sharing configurations can be derived from training dynamics rather than relying on manual design or expensive search, offering a scalable pathway for addressing interference in high-diversity multilingual settings.

\newpage
\section{Disclosure Statement}
This work is currently under review at an international peer-reviewed conference. The authors developed the core methodology and performed all experimental analyses independently. Generative AI tools (GitHub Copilot) were used exclusively for improving technical English clarity, correcting grammatical syntax, and enhancing narrative flow. All AI-generated suggestions were critically reviewed and validated by the authors. No conceptual, methodological, or substantive experimental work was generated by AI systems.

\bibliographystyle{IEEEtran}
\bibliography{mybib}

\clearpage
\onecolumn
\appendix
\section{Appendix A: Translation Quality Examples}

Representative translation examples across all four target languages are presented in Table~\ref{tab:translation_examples}. In this comparison, the \textbf{Baseline} refers to results obtained from the SeamlessM4T-Medium model after standard unified fine-tuning, while the proposed approach denotes results from our gradient-informed specialized architecture. The specialized model demonstrates improved semantic preservation, reduced hallucination, and more accurate handling of numerical expressions and named entities.

\begin{table*}[!ht]
\centering
\scriptsize
\renewcommand{\arraystretch}{1.2}
\setlength{\tabcolsep}{5pt}
\begin{tabularx}{\textwidth}{|l|l|X|X|X|}
\hline
\textbf{Lang} & \textbf{Audio ID} & \textbf{Reference} & \textbf{Baseline$^\dagger$} & \textbf{Proposed} \\
\hline
aeb & seg0002 & Ah, did you take her with you? & Uh, did you bring her with you? & Ah, did you take her with you? \\
\hline
aeb & seg0054 & Didn't you go out? & Did you go out with him? & Didn't you go out? \\
\hline
aeb & seg0087 & Ah, I thought it was the opposite. & Ah! Ah! What's wrong with you? & Ah, ah, it's the same thing. \\
\hline
aeb & seg0059 & Are you still thinking about when? & I'm going to do something else. & I started to think about it. \\
\hline
aeb & seg0083 & It's Eid, the next Sunday. & I mean, it's my birthday. & I'm coming on Sunday. \\
\hline
bem & elicit\_1 & the dog has something in its neck & The dog has something on its neck. & the dog has something in its neck \\
\hline
bem & b40\_elicit & How is this white woman and her two children who are playing looking like ? & How is this white woman with her three children playing? & How is this white mother who is with her three children who are playing looking? \\
\hline
bem & 5a5\_elicit & A man with a black suit and a white shirt is among the women. & A young man wearing a black suit and a white shirt is with three women. & A man wearing a black suit and a white shirt is among three women. \\
\hline
bem & ce8\_elicit & He is well braced and dressed in protective clothing. & He seems to be in a hurry and wearing protective gear. & He looks prepared to wear protective clothing. \\
\hline
bem & e94\_elicit & Yes and this man is sitted besides the river. & Yes, and the man is sitting on the side of the river. & Yes, and this man is seated at the side of the river. \\
\hline
est & 2291.539 & But this plus, that we are the smartest, it still comes as a matter of fact from the fact that you talk to these people. & That, but this plus, that how we're the smartest, it still comes down to the fact that you're talking to these people. & That, but this plus, that how we are now the smartest, it still comes as a matter of fact and that you talk to these people. \\
\hline
est & 1197.134 & Admitted that he secretly financed the Center Party in order to get a favorable decision from the Tallinn city government. & admitted that he secretly funded the Centre Party in order to get a favorable decision from the Tallinn City Government. & admitted that he secretly funded the Center Party in order to get a favorable decision from the Tallinn city government. \\
\hline
est & 4475.152 & So yes, eight terawatts to ten terawatt-hours of electricity per year. But to meet climate goals, there is still electrification. What we used to do with wood has to be replaced by electricity. & That yes, eight-point to ten-point electricity per year, but to fulfill climate goal everywhere, it is still electrification, what we used to do with wood, you have to replace it with electricity. & That yes, eight degrees to ten degrees of electricity per year, but in order to fulfill the climate goal all over Europe and the world, it is still electrification. What we used to do with wood has to be replaced by electricity. \\
\hline
est & 130.483 & Seven, three hundred two, four hundred sixty two is ringing, good day, you are on Vikerradio. & Seven, three hundred and two, four hundred and sixty-two sounds good day, you are on Vikerraadio. Good day. & Seven, three hundred and two, four hundred and sixty two is ringing, good day, you are on Vikerradio. Good day. \\
\hline
est & 922.975 & That he is bullied, but sometimes it can also happen that the child, who is in trouble with his own, starts to bully others. & that he is being bullied, but it may happen from time to time that the child who himself starts bullying others with his bullets. & That he is being bullied, but sometimes it can happen that the child who can bully others with his own hands. \\
\hline
gle & 18183766 & If we do this, a future full of possibilities lies ahead & If we do this, we will have a future full of possibilities & If we do this, a future full of possibilities awaits us \\
\hline
gle & 22316649 & I wish you all the best as you continue your important work. & I wish you every success as you continue with your important work. & I wish you all the best and you continue with your important work. \\
\hline
gle & 18183719 & Speech at an event to celebrate the tenth anniversary of Gaelscoil Cluainín & Speech at an event to celebrate the tenth anniversary of the Gaelscoil Cluainín & Speech at an event to celebrate the tenth anniversary of Gaelscoil Cluainín \\
\hline
gle & 36911346 & How are you, Taidhg & Where are you, Haig? & How are you, priest? \\
\hline
gle & 25317156 & The square root of three thousand four hundred and twenty six & The square root of three thousand four hundred and twenty-six & The square root of three thousand four hundred and twenty six \\
\hline
\end{tabularx}
\caption{Translation quality comparison. $^\dagger$Baseline refers to unified fine-tuned SeamlessM4T-Medium.}
\label{tab:translation_examples}
\end{table*}

\end{document}